\documentclass[sigconf]{acmart}

\AtBeginDocument{%
  }

\copyrightyear{2025}
\acmYear{2025}
\setcopyright{rightsretained}
\acmConference[KDD '25]{Proceedings of the 31st ACM SIGKDD Conference on
Knowledge Discovery and Data Mining V.1}{August 3--7, 2025}{Toronto, ON,
Canada}
\acmBooktitle{Proceedings of the 31st ACM SIGKDD Conference on Knowledge
Discovery and Data Mining V.1 (KDD '25), August 3--7, 2025, Toronto, ON,
Canada}
\acmDOI{10.1145/3690624.3709292}
\acmISBN{979-8-4007-1245-6/25/08}

\usepackage{graphicx}
\usepackage{booktabs} 
\usepackage{hyperref}

\usepackage{amsmath}
\usepackage{amsthm}
\usepackage{algorithm}
\usepackage{algorithmic}
\usepackage{mathtools}
\usepackage[capitalize,noabbrev]{cleveref}
\usepackage{graphicx}
\usepackage{booktabs}
\usepackage{wrapfig}
\usepackage{lipsum}
\usepackage{booktabs}
\usepackage{xspace}
\usepackage{enumitem}
\usepackage{hyperref}
\usepackage{caption}
\usepackage{subcaption}
\usepackage{subcaption}
\usepackage{orcidlink}
\usepackage{xcolor,colortbl}
\usepackage{natbib}
\newcommand{\method}{LResNet\xspace}
\usepackage{orcidlink}
\newcommand{\x}{\mathbf{x}}
\newcommand{\y}{\mathbf{y}}
\newcommand{\z}{\mathbf{z}}
\newcommand{\R}{\mathbb{R}}

\usepackage[english]{babel}
\usepackage{hyperref}       

\renewenvironment{proof}{\begin{newproof}}{\end{newproof}\qed}

\theoremstyle{plain}
\newtheorem{theorem}{Theorem}[section]
\newtheorem{proposition}[theorem]{Proposition}
\newtheorem{lemma}[theorem]{Lemma}

\theoremstyle{definition}

\theoremstyle{remark}

\begin{document}

\title{Lorentzian Residual Neural Networks}

\author{Neil He}
\email{neil.he@yale.edu}
\orcid{0009-0008-3193-2448}
\affiliation{%
  \institution{Yale University}
  \city{New Haven}
  \country{United States}
}
\author{Menglin Yang}
\email{menglin.yang@yale.edu}
\orcid{0000-0003-2510-5282}
\affiliation{%
  \institution{Yale University}
  \city{New Haven}
  \country{United States}
}
\author{Rex Ying}
\email{rex.ying@yale.edu}
\orcid{0000-0002-5856-5229}
\affiliation{%
  \institution{Yale University}
  \city{New Haven}
  \country{United States}
}

\thanks{Corresponding author: Menglin Yang \\
\color{blue!50!black}{\url{https://github.com/Graph-and-Geometric-Learning/LResNet}}}

\begin{abstract}
Hyperbolic neural networks have emerged as a powerful tool for modeling hierarchical data structures prevalent in real-world datasets. 
Notably, residual connections, which facilitate the direct flow of information across layers, have been instrumental in the success of deep neural networks. 
However, current methods for constructing hyperbolic residual networks suffer from limitations such as increased model complexity, numerical instability, and errors due to multiple mappings to and from the tangent space. To address these limitations, we introduce \method, a novel Lorentzian residual neural network based on the \textit{weighted Lorentzian centroid in the Lorentz model of hyperbolic geometry}. Our method enables the efficient integration of residual connections in Lorentz hyperbolic neural networks while preserving their hierarchical representation capabilities. 
We demonstrate that our method can theoretically derive previous methods while offering improved stability, efficiency, and effectiveness. 
Extensive experiments on both graph and vision tasks showcase the superior performance and robustness of our method compared to state-of-the-art Euclidean and hyperbolic alternatives. Our findings highlight the potential of \method for building more expressive neural networks in hyperbolic embedding space as a generally applicable method to multiple architectures, including CNNs, GNNs, and graph Transformers.

\end{abstract}


\renewcommand{\shortauthors}{Neil He et al.}


\maketitle

\section{Introduction}
\label{sec:intro}
In recent years, exploration of neural network architectures beyond traditional Euclidean space has opened up new frontiers in machine learning research~\cite{peng2021hyperbolic,mettes2023hyperbolic,yang2022hyperbolicsurvey}. Among these, hyperbolic neural networks~\cite{HNN,HNN++,van2023poincar,hgcn2019,liu2019HGNN,Bdeir2024fully,chen2021fully} have gained significant attention due to their inherent capabilities to model data with complex hierarchical structures. Hyperbolic spaces, characterized by their constant negative curvature that allows for exponential growth of volume, naturally align with the geometric properties of tree-like data, offering a more fitting representation than their Euclidean counterparts that suffer from embedding distortions~\cite{2010hyperbolic,sarkar2011low}. This alignment has the potential to enhance learning efficiency and improve representation learning for a wide range of applications, from graph-based data analysis~\cite{faqeeh2018characterizing,yang2022htgn,sun2021hgcf,yang2022hrcf,yang2022hicf,liu2019HGNN,hgcn2019} to image understanding~\cite{HistopathologyHyperbolic,chen2023hyperbolic,weng2021unsupervised,ermolov2022hyperbolic}.

One of the core elements of the modern deep learning framework is the residual connection~\cite{he2016deep}, a powerful mechanism that has revolutionized the development of deep neural networks. 
By allowing layers to learn modifications to the identity mapping rather than complete transformations, which addresses issues such as the gradient vanishing and graph over-smoothing problems, residual connections facilitate the training of substantially deeper networks and are widely used in many model architectures, including CNNs, GNNs, transformers, and diffusion models~\cite{he2016deep, gat2018, vaswani2017attention,song2021scorebased}.
 However, the application of residual connections within Euclidean spaces is not directly transferable to the complex geometry of hyperbolic spaces. This is primarily due to the curvature of hyperbolic space, where direct addition often violates geometric constraints. In the Poincar{\'e} ball model, the sum could exceed the boundary of the ball~\cite{ungar2008gyrovector}. Similarly, in the Lorentz model, it could cause points to deviate from the hyperboloid manifold~\cite{ramsay2013introduction}.

\begin{figure*} 
    \centering
   \begin{subfigure}{0.28\textwidth}
        \centering
       \includegraphics[width=\linewidth]{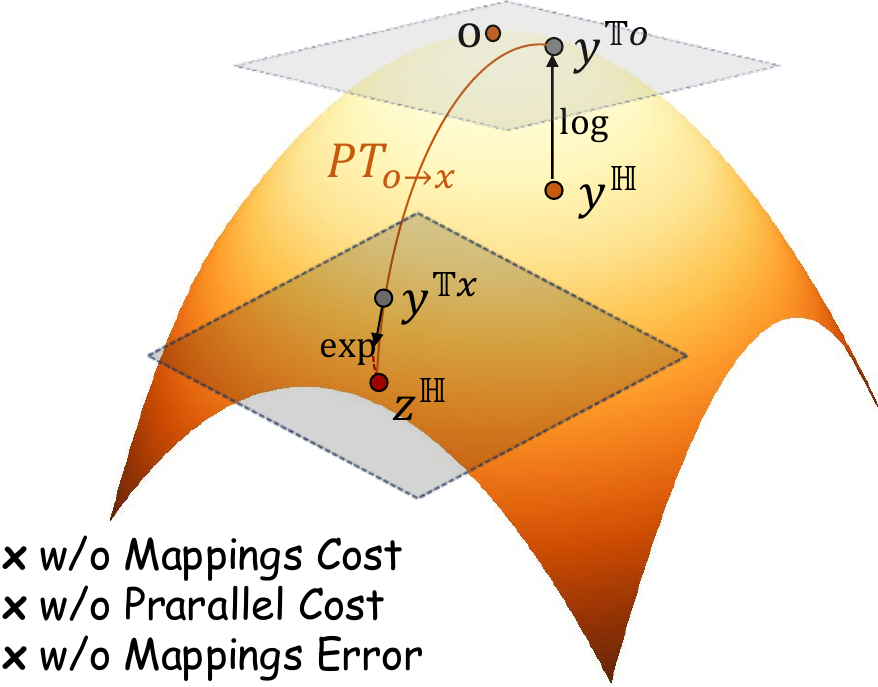}
       \caption{Parallel Transport Based Method}
       \label{fig:mobius}
   \end{subfigure}
\hspace{10pt} 
   \begin{subfigure}{0.28\textwidth}
        \centering   
       \includegraphics[width=\linewidth]{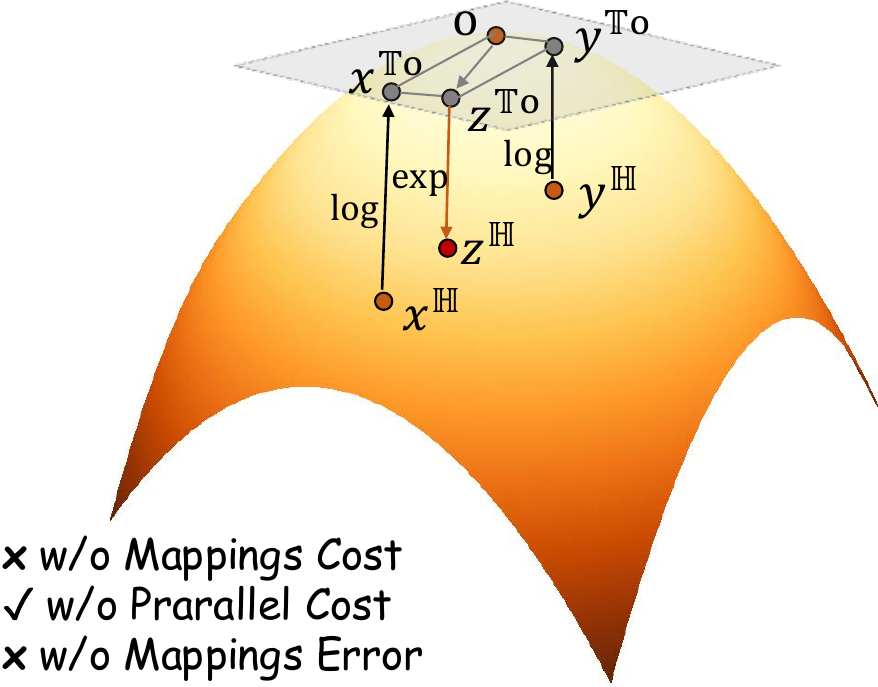}
       \caption{Tangent Space Based Method}
       \label{fig:tangent}
   \end{subfigure}
\hspace{10pt} 
   \begin{subfigure}{0.28\textwidth}
        \centering
       \includegraphics[width=\linewidth]{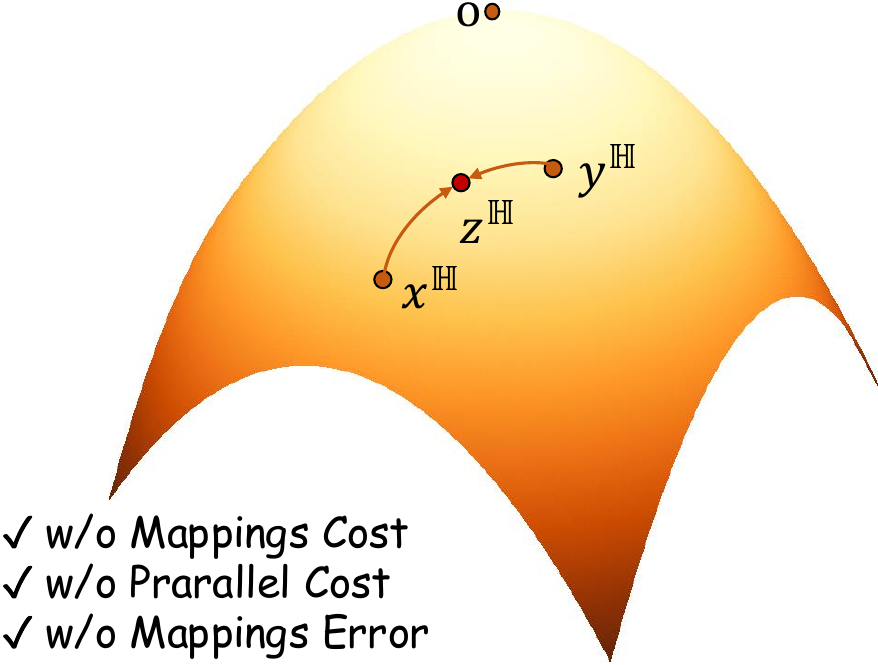}
       \caption{\method (Ours)}
       \label{fig:midpoint}
   \end{subfigure}
   \caption{Visualization of hyperbolic residual connection methods. From left to right: (a) Parallel transport-based method, (b) Tangent space-based method, and (c) The proposed \method. Points with superscript $\mathbb{H}$ and $\mathbb{T}$ indicate their presence in hyperbolic space and tangent space, respectively.  In each subfigure, $\z^\mathbb{H}$ represents the sum of points $\x^\mathbb{H},\y^\mathbb{H}\in\mathbb{H}$.
   PT denotes parallel Transport, 
   and \textbf{log} and \textbf{exp} denotes the logarithmic and exponential mappings respectively. Our proposed method \method overcomes limitations L(i, ii, iii, iv) by eliminating mappings and parallel transport (whose absence is shown via $\checkmark$), where the other two methods depend on as least one (shown via $\times$).
   } 
   \label{fig:visuals}
\end{figure*}
\textbf{Existing works and their limitations.} \label{prev} Several previous works have explored implementing residual connections in hyperbolic space. Poincar\'e ResNet~\cite{van2023poincar} proposes projecting the hyperbolic embeddings to the tangent space at a fixed point, parallel transporting it to the tangent space at the position of interest, and then utilizing the exponential map to map them back to the hyperbolic space as shown in ~\cref{fig:mobius}. Another similar work, Riemannian ResNet~\cite{katsman2023riemannian}, although not confined just to hyperbolic space, employs similar concepts for an analogous method. As a method previously used as hyperbolic addition for aggregation, HGCN~\cite{hgcn2019} and LGCN~\cite{lgcn} propose first mapping to the tangent space of the origin for addition, and then projecting the sum back into hyperbolic space, as shown in ~\cref{fig:tangent}. HCNN~\cite{Bdeir2024fully} proposes adding the Lorentzian space-like component of the hyperbolic embeddings and then computing the time-like component afterward. Each of these previous methods suffers from at least one of the limitations listed below:
\begin{enumerate}[label=(\roman*)]
\renewcommand{\labelitemi}{$$}
    \item {Computationally Expensive.} The parallel transport and tangent space addition methods involve complex mappings to and from the tangent space, significantly increasing computational complexity (see \cref{time_analysis} for runtime comparison).
    
    \item {Non-Commutativity}. The Euclidean residual connection proposed in~\cite{he2016deep} satisfies the commutative property $\mathbf{x} + f(\mathbf{x}) = f(\mathbf{x}) + \mathbf{x}$, while the parallel transport-based residual connection is non-commutative. This greatly restricting the expressiveness and flexibility of the model (see~\cref{non-commute}).

    \item {Numerical Instability}. The parallel transport and tangent space addition methods are prone to numerical instability. For curvature -1, the geodesic distance in the $n$ dimensional Poincar\'e ball model $\mathcal{P}^n$ and the logarithmic map in the $n$ dimensional Lorentz model $\mathcal{L}^n$ are defined by the following formulas, respectively:
        \begin{equation}
        \begin{aligned}
            d(\mathbf{x}, \mathbf{y})&=\cosh ^{-1}\left(1+2 \frac{\|\mathbf{x}-\mathbf{y}\|^2}{\left(1-\|\mathbf{x}\|^2\right)\left(1-\|\mathbf{y}\|^2\right)}\right), \\
            \log _{\mathbf{u}}(\mathbf{v})&=\frac{\cosh ^{-1}(\alpha)}{\sqrt{\alpha^2-1}}(\mathbf{v}-\alpha \mathbf{u}),
            \end{aligned}
        \end{equation}
where $\mathbf{x}, \mathbf{y}\in \mathcal{P}^n$, $\mathbf{u}, \mathbf{v}\in \mathcal{L}^n$,
$\alpha=\mathbf{u}_0 \mathbf{v}_0-\sum_{i=1}^n \mathbf{u}_i \mathbf{v}_i$ is the Lorentzian inner product of $\mathbf{u}, \mathbf{v}$. For Poincar\'e distance, when $\mathbf{x}, \mathbf{y}$ are near the boundary, the floating point representation of $\|\mathbf{x}\|^2,\|\mathbf{y}\|^2$ approaches one and the denominator of the geodesic distance becomes zero, making parallel transport numerically instable as the operation depends on dividing by distance. For the Lorentz model, if $\mathbf{u}=\mathbf{v}$ and $\mathbf{u}$ contains large coordinate values, $\alpha$ becomes less than one, which results in NaN because the domain of $\cosh ^{-1}(x)$ is $x \geq 1$.

\item {Mapping Error}. The parallel transport and tangent space methods require mapping a point in hyperbolic space to the tangent space, typically using the origin as the reference point for efficient computation. However, this introduces mapping errors for points not at the origin, especially for those far from it~\cite{yu2019numerically}.

\item {Lack of Geometric Meaning.} The Lorentzian space-like dimension addition method from HCNN~\cite{Bdeir2024fully} lacks a clear geometric interpretation within hyperbolic geometry, providing no motivation and justification as to why it works.
\end{enumerate}
In particular, the parallel transport method proposed by Poincar\'e ResNet~\cite{van2023poincar} and Riemmanian ResNet~\cite{katsman2023riemannian} suffer from (i), (ii), (iii), (iv). The tangent space addition method used in HGCN~\cite{hgcn2019} and LGCN~\cite{lgcn}, suffers from (i), (ii), (iv). The space addition method from HCNN~\cite{Bdeir2024fully} suffers from (v). In \cref{fig:visuals}, we show the visualization of \method and two of the previous methods: the parallel transport method and tangent space. As demonstrated, these two methods suffer from the need for multiple mappings to and from the tangent space while our method operates entirely on the hyperbolic manifold. The space addition method is not shown as it does not have geometric interpretability.

\textbf{Proposed method. } 
To overcome the above limitations, we propose \textbf{L}orentzian \textbf{Res}idual \textbf{Net}works (\method), a residual neural network utilizing the weighted Lorentzian centroid. Unlike existing methods that use parallel transport and tangent space addition, our approach normalizes the Euclidean weighted sum to directly project the output into the Lorentz model. Consequently, we avoid the need to map between tangent and hyperbolic spaces, operating directly on the manifold. This approach addresses the limitations (i), (iii), and (iv), and ensures the commutativity of addition, thereby resolving limitation (ii). Unlike HCNN~\cite{Bdeir2024fully}, our method has geometric interpretations and alleviates limitation (v), both in the form discussed in \cref{main_prop} and as the centroid w.r.t. to the Lorentzian distance~\cite{law2019lorentzian}. Theoretically, we demonstrate that the proposed method can derive all of the discussed previous methods, offering general applicability while ensuring stability (see \cref{stability lemma}). Experimentally, our method achieves superior performance across multiple graph and computer vision task datasets while being effective in addressing the graph over-smoothing problem. 

We demonstrate the versatility of \method with adaptations to several model architectures, achieving significant performance improvements: (1) GNNs, with up to 10.9\% improvement over previous hyperbolic residual connection methods; (2) graph Transformers, with up to 2.5\% improvement; and (3) CNNs for vision tasks, with up to 0.5\% improvement and more robustness. Additionally, our method achieves over 2,000 times faster computation when compared to the parallel transport and tangent space method. It should be noted that \method can be applied to \textbf{any} hyperbolic neural network that operates on the Lorentz hyperboloid. Additionally, HNN++\cite{HNN++} has proven that previous midpoint computations proposed for the Beltrami-Klein model and Poincar{\'e} ball model \cite{ungar2008gyrovector} are equivalent to the Lorentzian centroid~\cite{law2019lorentzian} under isometric projections. As \method is based on a generalization of the Lorentzian centroid, it can potentially be extended to other hyperbolic formulations via these mappings.

\textbf{Contributions.} The main contributions of this work can be summarized as follows:
(1) \textbf{Introduction of \method}. We introduce \method, a numerically stable foundational residual connection method for hyperbolic neural networks that does not rely on the tangent space and parallel transport, thus being more efficient and effective than previous methods.
(2) \textbf{Broad Applicability}. We successfully apply the proposed method \method across various domains and tasks, including computer vision and graph tasks, and across multiple architectures such as CNN, GNN, and graph Transformer.
(3) \textbf{Theoretical Insights}. We theoretically analyze the limitations of existing methods and show that the proposed method encompasses previous methods, greatly enhancing the understanding of hyperbolic residual networks.

\section{Related Works}\label{related works}

\textbf{Hyperbolic representation and deep learning. }
Hyperbolic spaces have garnered extensive attention in recent years for representation learning and deep learning~\cite{peng2021hyperbolic,mettes2023hyperbolic,yang2022hyperbolicsurvey}. A defining geometric characteristic of hyperbolic spaces is their negative curvature, which results in an exponential growth of volume with distance from the origin. This property closely mirrors the structure of tree-like data, where the number of child nodes increases exponentially~\cite{2010hyperbolic}. Consequently, hyperbolic representation learning provides a strong geometric prior for hierarchical structures, tree-like data, and power-law distributed information. 
Significant advancements have been achieved in various domains such as WordNet~\cite{nickel2017poincare,nickel2018learning}, graphs~\cite{hgcn2019,liu2019HGNN,zhang2021hyperbolic,yang2023hyperbolic,yang2023kappahgcn,yang2024hypformer}, social networks~\cite{yang2021discrete,yang2022htgn}, and recommendation systems~\cite{yang2022hrcf,yang2022hicf,chen2021modeling,sun2021hgcf} utilizing hyperbolic representations. Moreover, hyperbolic deep learning has demonstrated impressive performance in image-related tasks, including image embedding~\cite{khrulkov2020hyperbolic,guo2022clipped,ermolov2022hyperbolic} and segmentation~\cite{atigh2022hyperbolic,weng2021unsupervised,chen2023hyperbolic}, offering new insights into deep learning paradigms, such as the interpretation of norms as reflections of uncertainty.
Within the neural network domain, HNN~\cite{HNN} presented the pioneering form of hyperbolic neural networks, defining fundamental hyperbolic transformations, classifiers, and hyperbolic multiclass logistic regression (MLR). HNN++~\cite{HNN++} further reduced the parameter count of hyperbolic MLR and made advancements in fully connected layers, the splitting and concatenation of coordinates, convolutional layers, and attention mechanisms within hyperbolic space. Recent studies, such as HyboNet~\cite{chen2021fully} and Hypformer~\cite{yang2024hypformer} propose frameworks that construct neural networks entirely within hyperbolic space, contrasting with existing models that partially operate in Euclidean tangent spaces.

\textbf{Residual neural networks and their hyperbolic adaptations.} 
Residual connections are a fundamental module in modern neural networks~\cite{he2016deep}, addressing the vanishing gradient problem directly and enabling the training of significantly deeper networks. By supporting identity mapping through skip connections, they enhance gradient flow across layers, thereby improving training efficiency and model performance across a diverse set of tasks.

Several works have explored the adaptation of residual connections to hyperbolic spaces. Poincar{\'e} ResNet~\cite{van2023poincar} leverages the concept of parallel transport to map points in hyperbolic space to the tangent space at a reference point and then, via parallel transport, map to the tangent space of a new point, and subsequently map them back using the exponential map. Riemannian ResNet~\cite{katsman2023riemannian} proposes an analogous method to perform this operation. HCNN~\cite{Bdeir2024fully} proposes to sum the Lorentzian space-like components and uses post-summation processing on the time component.

However, these approaches have several limitations, as discussed in the introduction. This work aims to address these challenges.

\section{Preliminaries}\label{preliminaries}
This section provides an overview of the fundamental concepts in hyperbolic geometry, focusing on the Lorentz model. Hyperbolic space can be formulated using several models, including the Poincar{\'e} ball model~\cite{nickel2017poincare}, the Lorentz (Hyperboloid) model~\cite{nickel2018learning}, and the Klein model~\cite{gulcehre2019hyperbolicAT}. These models are isometric, meaning that points in one model can be smoothly mapped to another while preserving distances, angles, and geodesics~\cite{ramsay2013introduction}.

\textbf{Lorentz model. } 
An $n$-dimensional Lorentz model is
a Riemannian manifold $(\mathcal{L}^n, \mathfrak{g}_n^K)$ equipped with the Riemannian metric tensor $\mathfrak{g}_n^K = \mathrm{diag}(-1, 1, \ldots, 1)$ and defined by a constant negative curvature $K<0$, denoted as $\mathbb{L}^{K,n}$. Each point $\x\in\mathbb{L}^{K,n}$ has the parametrized form $[x_t, \x_s]^T$ where $x_t\in\R$ is called the time-like component and $\x_s\in\R^{n}$ is called the space-like component. $\mathbb{L}^{K,n}$ is equipped with the \textit{Lorentzian inner product}. For points $\x,\y\in\mathbb{L}^{K,n}$, their inner product $\langle\x,\y\rangle_\mathcal{L}$ is given by 
\begin{align}
    \langle\x,\y\rangle_{\mathcal{L}} &= -x_ty_t + \x_s^T\y_s = \x^T\mathfrak{g}_n^K\y,
\end{align}
with $|\|\x\||_\mathcal{L}\coloneq\sqrt{|\langle \x, \x\rangle_\mathcal{L}|}$ being the Lorentzian norm. Formally, $\mathcal{L}^n$ is the point set \begin{displaymath}
    \mathcal{L}^n = \{\x\in\R^{n+1}: \langle\x,\x\rangle_\mathcal{L} = 1/K, x_t>0\}.
\end{displaymath}
The origin $\mathbf{o}\in\mathbb{L}^{K,n}$ is the point $[\sqrt{-1/K}, 0,\ldots,0]^T$.

\textbf{Tangent space.}
The tangent space at a point $\x\in\mathbb{L}^{K,n}$ is set of points orthogonal to $\x$, defined as \begin{displaymath}
    \mathcal{T}_\mathbf{x}\mathbb{L}^{K,n} = \{\y\in\R^{n+1}: \langle\x,\y\rangle_{\mathcal{L}} =0 \}.
\end{displaymath}
Notably, the tangent space is isometric to Euclidean space.

\textbf{Exponential and logarithmic maps. }
For each point $\x\in\mathbb{L}^{K,n}$, the exponential map $\exp_\x^K:\mathcal{T}_\x\mathbb{L}^{K,n}\to \mathbb{L}^{K,n}$ and the logarithmic map $\log_\x^K:\mathbb{L}^{K,n}\to \mathcal{T}_\mathbf{x}\mathbb{L}^{K,n}$ at $\mathbf{x}$ are given by
\begin{equation}
        \exp_\x^K(\y) = \cosh(\alpha)\x + \frac{\sinh(\alpha)}{\alpha}\y,
 \alpha = \sqrt{-K\langle\x,\y\rangle_\mathcal{L}},
\end{equation}
\begin{equation}
   \log_\x^K(\x)= \frac{\cosh^{-1}(\beta)}{\sqrt{\beta^2-1}}(\y-\beta\x),
     \beta = K\langle\x,\y\rangle_\mathcal{L}.
\end{equation}

\textbf{Parallel transport.}
    Parallel transport is a generalization of translation to hyperbolic geometry mapping a point $\z\in\mathcal{T}_\x\mathbb{L}^{K,n}$ to a point in $\mathcal{T}_\y\mathbb{L}^{K,n}$ via \begin{equation*}
        \mathbf{P}_{\x\to\y}(\z) = \z + \frac{\langle\y,\z\rangle_\mathcal{L}}{-1/K-\langle\x,\y\rangle_\mathcal{L}}(\x+\y).
\end{equation*}

\textbf{Hyperbolic addition methods.}\label{prev_add_methods}
Several methods have been proposed for performing addition in hyperbolic space.
The parallel transport method of M\"obius addition from~\cite{hgcn2019,HNN} is given by \begin{equation}\label{pt}
    \x\oplus_P\y = \exp_{\x}^K\circ\mathbf{P}_{\mathbf{o}\to\x}\circ\log_{\mathbf{o}}^K(\y).
\end{equation}

Another approach is the tangent space addition method used for aggregation in \cite{hgcn2019}, given as:
\begin{equation}\label{ts}
    \x\oplus_T\y = \exp_{\mathbf{o}}^K\left(w_x\log_{\mathbf{o}}^K(\x) +w_y\log_{\mathbf{o}}^K(\y) \right),
\end{equation}
where $w_x,w_y>0$ are weights. A third approach is the space-like dimension addition method from \cite{Bdeir2024fully} given by \begin{equation}\label{sa}
    \x\oplus_S\y = \left[\sqrt{||\x_s+\y_s||^2 - 1/K}, \mathbf{\x_s+\y}_s\right]^T,
\end{equation}
where $\x_s, \y_s$ denote the space-like dimension of $\x,\y$.

\section{Methodology} \label{methods}
In this section, we introduce our proposed method for Lorentzian residual connection, based on a generalization of the Lorentzian centroid~\cite{law2019lorentzian}, and analyze its theoretical properties, including numerical stability and representation power. We also propose the use of an optional scaling method that helps control the norm of the output from the residual layer.

\subsection{Lorentzian residual connection.}\label{method}
In a standard Euclidean residual block, let $\x$ and $f(\x)$ represent the input and output from a neural network layer or series of layers. The residual connection is expressed as $\x + f(\x)$, or more generally, as $\alpha \x + \beta f(\x)$, where $\alpha$ and $\beta$ are scalar weights.

Given vectors $\x,f(\x)\in\mathbb{L}^{K,n}$, the Lorentzian residual connection is defined as follows:
\begin{equation} \label{midpoint formula}
\mathbf{x}\oplus_\mathcal{L} f(\mathbf{x}) := \frac{w_x\mathbf{x} + w_y f(\mathbf{x})}{\sqrt{-K} |\| w_x\mathbf{x} + w_y f(\mathbf{x}) \|_{\mathcal{L}}|},
\end{equation}
where  $\left|\left\|\cdot\right\|\right|_{\mathcal{L}}=\sqrt{|\langle\cdot\rangle_\mathcal{L}|}$ is the Lorentzian norm and $w_x,w_y>0$ are weights that can be learned or fixed. This formulation projects the Euclidean weighted sum directly onto $\mathbb{L}^{K,n}$ using the normalizing denominator to ensure it lies in the Lorentz hyperboloid.

Following the general form of the residual connection in the Euclidean case, we can reformulate \cref{midpoint formula} as a weighted sum of Lorentzian hyperbolic vector, given by \begin{equation}\label{lrn formula}
    \mathbf{x}\oplus_\mathcal{L} f(\mathbf{x}) := \alpha_{w_x,w_y} \mathbf{x} + \beta_{w_x,w_y} f(\mathbf{x}),
\end{equation}
where \[\alpha_{w_x,w_y} = {w_x}/{\sqrt{-K} |\| w_x\mathbf{x} + w_y f(\mathbf{x}) \|_{\mathcal{L}}|}\] \[\beta_{w_x,w_y} = {w_y}/{\sqrt{-K} |\| w_x\mathbf{x} + w_y f(\mathbf{x}) \|_{\mathcal{L}}|}\] 
are normalized weights. 

\textbf{Lorentz residual network (\method) } The core component of \method is the Lorentzian residual block, which consists of a hyperbolic layer followed by a Lorentzian residual connection. The hyperbolic layer can be any type of layer adapted to the Lorentz model, such as hyperbolic fully-connected layers~\cite{chen2021fully}, hyperbolic convolutional layers~\cite{Bdeir2024fully}, or hyperbolic attention mechanisms~\cite{gulcehre2019hyperbolicAT}. In \cref{algo}, we demonstrate \method applied to a classification network as an example. However, this method is not confined to this particular scenario. To ensure that the weights are constraints to the feasible domain, in practice, we fix $w_x$ to be a positive constant and we take the absolute value of $w_y$. Please see \cref{stability lemma} and the accompanying discussion for more details on the feasible domain.

Next, we study the theoretical aspects of \method. We show that \method is numerically stable in \cref{stability lemma} and that \method generalizes previous methods in \cref{main_prop}. We also show that \method is a valid hyperbolic operation and that it satisfies the commutativity property. The complete proofs can be found in \cref{proofs}.

\textbf{Numerical stability.}
TWe show that the weights $\alpha_{w_x,w_y}$ and $\beta_{w_x,w_y}$ can always be selected to ensure that \method is numerically stable. Note that the only possible source of numerical instability is from the division, hence it suffices to ensure the dominator is lower bounded by some constant and does not approach zero, which we show via the following lemma.
\begin{lemma}\label{stability lemma}
    $\sqrt{-K}|\|w_x\x+w_y\y\|_\mathcal{L}|>\sqrt{w_x^2 + w_y^2}\,$ for any $\,\x,\y\in\mathbb{L}^{K,n}$ and $(w_x,w_y)\in\R^+\times\R^+\setminus\{(0,0)\}$, where $\R^+$ denotes the set of non-negative real numbers.
\end{lemma}
Since the normalizing denominator divides out the ratio $w_x/w_y$, the output of \method is unchanged when the ratio is preserved. Thus we can fix $w_x$ to be a positive constant. In the special case of $w_x = 1$, we obtain $||w_x\x + w_y\y||_\mathcal{L}>1/\sqrt{-K}>0$ from the lemma. This lemma ensures that the denominator can always be lower-bounded by a positive constant of our choosing (in this case, $1/\sqrt{-K}$). Thus we never risk dividing by values close to zero, making $\method$ \textbf{numerically stable}. Note that $w_y$ is still allowed to take on any arbitrary non-negative value, which allows for any arbitrary ratio $w_x/w_y$. Thus fixing $w_x = 1$ does not at all restrict the values of the output of \cref{lrn formula} and any discussion of using arbitrary $w_x$ values remain applicable.

\textbf{Validity of \method. } Here we show that \method is a valid hyperbolic operation.
To see that \cref{lrn formula} indeed maps to hyperbolic space, first note that for any $\x, f(\x)\in\mathbb{L}^{K, n}$, we can compute \begin{align*}
    &\left\langle\alpha_{w_x,w_y} \mathbf{x} + \beta_{w_x,w_y} f(\mathbf{x}), \alpha_{w_x,w_y} \mathbf{x} + \beta_{w_x,w_y} f(\mathbf{x})\right\rangle_\mathcal{L}\\
    &= \frac{\langle w_x\x + w_yf(\x), w_x\x + w_yf(\x)\rangle_\mathcal{L}}{-K |\| w_x\x + w_yf(\x), w_x\x + w_yf(\x)\|_\mathcal{L}|}\\
    &= \pm 1/K.
\end{align*}
Given that $\alpha_{w_x,w_y} \mathbf{x} + \beta_{w_x,w_y} f(\mathbf{x})$ is a positive time-like vector, its Lorentzian inner product with itself must be negative, specifically it must be $-1/K$ from above. Therefore $\alpha_{w_x,w_y} \mathbf{x} + \beta_{w_x,w_y} f(\mathbf{x})\in\mathbb{L}^{K,n}$, confirming that \method is a valid hyperbolic operation. 

\textbf{Relation to previous methods. }
Our approach 
can theoretically derive the geometric meaning of the previous methods mentioned in \cref{preliminaries}, based on the following results in Proposition~\ref{main_prop}. 
\begin{proposition}
\label{main_prop}
    Let $\mathbf{z}$ be the output of one of the parallel transport methods in \cref{pt}, the tangent space method in \cref{ts}, or the space addition method in \cref{sa}. Then there exists weights $w_x,w_y\in \mathbb{R}^+$ such that the point $\mathbf{m} = \alpha_{w_x,w_y}\x + \beta_{w_x,w_y}\y$ lies on the geodesic from $\mathbf{o}$ to $\mathbf{z}$.
\end{proposition}

Then by carefully selecting weights (or using trainable weights) in \method, we can derive previous methods using our method. This shows that \method has at least the representative power of any of the previous methods, ensuring its expressiveness.

\textbf{Commutativity.} \method is commutative as it is a generalization of a weighted sum and $w_x\x + w_y\y = w_y\y+w_x\x$. For completeness and rigor, we also include the following theorem that demonstrates the non-commutativity of the parallel transport method. 
\begin{theorem}
\label{theorem:non-commutative}
    Let $\x,\y\in\mathbb{L}^{K,n}$ be points such that $\x_i = \y_i$ for $i\ne n+1$ and $\x_{n+1} = -\y_{n+1}$. Let $\z = \x\oplus_P\y$ be the output of the parallel transport method, and let $\z' = \y\oplus_P \x$ be the output in the other direction. Then $\z_{n+1} = -\z'_{n+1}$. 
\end{theorem}
The two directions of the parallel transport addition in the above case are reflected over an entire axis, giving theoretical motivation for its inflexibility and results shown in \cref{analysis}. For instance, when $n = 2$, $K = -1$, $\x = [3, 2, -2]^T$ and $\y = [3, 2, 2]^T$, we have $\x \oplus_P \y = [9, 8, -4]^T$ and $\y \oplus_P \x = [9, 8, 4]^T$.

\textbf{Computational Complexity. } To compute the weighted sum for \method in \cref{lrn formula}, we calculate the Lorentzian norm in the denominator of the weights $\alpha_{w_x, w_y}$ and $\beta_{w_x, w_y}$, which takes $O(n)$ time. Afterward, the remaining computation is a straightforward Euclidean addition, allowing \method to be computed in $O(n)$ time, comparable to the computational efficiency of the Euclidean case.

\begin{algorithm}[tb]
\caption{Lorentz Residual Network (\method)}\label{algo}
\begin{algorithmic}[1]
\REQUIRE Input data $\x \in \mathbb{L}^{K,n}$, target labels $y$, learning rate $\eta$, number of layers $L$, loss function $\ell$, initial weights $w_x^{(l)}$ and $w_y^{(l)}$ for $l = 1, \ldots, L$, hyperbolic layers $f^{(l)}:\mathbb{L}^{K,n}\to\mathbb{L}^{K,n}$ for $l = 1, \ldots, L$
\FOR{each epoch}
    \FOR{each batch of data $\x_b$}
        \STATE $\mathbf{h}^{(0)} \gets \x_b$ {{// \color{black!50!blue}{Initialization}} }
        \FOR{$l \gets 1$ to $L$}
            \STATE $\mathbf{z}^{(l)} \gets f^{(l)}(\mathbf{h}^{(l-1)})$ \quad {{// \color{black!50!blue}Hyperbolic layers}}
            \STATE $\alpha^{(l)} \gets \frac{w_x^{(l)}}{\sqrt{-K} |\| w_x\mathbf{x} + w_y f(\mathbf{x}) \|_{\mathcal{L}}|}$
            \STATE $\beta^{(l)} \gets \frac{w_y^{(l)}}{\sqrt{-K} |\| w_x\mathbf{x} + w_y f(\mathbf{x}) \|_{\mathcal{L}}|}$
            \STATE $\mathbf{h}^{(l)} \gets \alpha^{(l)} \mathbf{h}^{(l-1)} + \beta^{(l)} \mathbf{z}^{(l)}$  {{// \color{black!50!blue}\method}}
        \ENDFOR
        \STATE $\hat{\mathbf{y}} \gets \text{softmax}(\mathbf{h}^{(L)})$ \quad {// \color{black!50!blue}Output prediction}
        \STATE $\mathcal{J} \gets \ell(\hat{\mathbf{y}}, \mathbf{y})$ \quad {// \color{black!50!blue}Compute loss}
        \STATE Update weights $w_x^{(l)}$ and $w_y^{(l)}$ for $l = 1, \ldots, L$ using optimizer with $\mathcal{J}$ and $\eta$
    \ENDFOR
\ENDFOR
\end{algorithmic}
\end{algorithm}
\subsection{Advantages over previous approaches} 
\method overcomes all of the previous methods' limitations discussed in~\cref{sec:intro}. In this section, we summarize the advantages below.
\begin{enumerate}[label=(\roman*)]
\renewcommand{\labelitemi}{$$}
\item \textbf{Efficiency.} Unlike previous methods that involve multiple mappings between hyperbolic and tangent spaces, \method is a simple weighted sum, making it significantly more efficient. \cref{analysis} provides a more detailed runtime analysis in \cref{time table}, where \method achieves over \textit{2000 times} faster computation than both the parallel transport and tangent space method in the same setting.
\item \textbf{Commutativity}. \method is commutative. In the \cref{method} and \cref{analysis}, we discuss the effects of the non-commutativity on the parallel transport method~\cite{hgcn2019}, showing that it is unpredictable which direction of addition achieves better performance.
\item \textbf{Numerical Stability.} The result from \cref{stability lemma} ensures that \method is computationally stable, avoiding the computational instability issues mentioned in the introduction. By eliminating the mapping between tangent and hyperbolic spaces, \method avoids mapping errors for points far away from the origin~\cite{yu2019numerically}. 

\item \textbf{Geometric Interpretation.} \method has a geometric interpretation, with the ability to theoretically achieve previous methods (Proposition~\ref{main_prop}). By carefully selecting weights, \method is able to achieve the geometric meaning of previous methods by ensuring the outputs lie on the same geodesic from the origin. This ensures the representation power of \method and provides theoretical motivation for \method as opposed to the space addition method~\cite{Bdeir2024fully}.
\end{enumerate}

\subsection{Optional scaling}
Very often the Euclidean norm of the hyperbolic embedding has important implications in model performance. For example, in image classification tasks, the norm tends to be positively correlated with classification confidence~\cite{ghadimi2021hyperbolic}. However, previous work~\cite{law2019lorentzian} has shown that the unweighted Lorentzian centroid (a specific case of \method) can get very close to the origin in manifolds with low curvature, resulting in a small Euclidean norm. Conversely, excessively large norms can lead to mapping errors~\cite{dooley1993harmonic}. To help keep the norm of the output within reasonable ranges, we propose to use an optional scaling method after the \method computation when training in low-curvature manifolds. For a hyperbolic vector $\mathbf{m}\in\mathbb{L}^{K,n}$, the scaled value is 
\begin{equation}\label{scaling formula}
    \begin{bmatrix}
        \sqrt{||\gamma\cdot\mathbf{m}_s||^2 - \frac{1}{K}}\\ \gamma \cdot \mathbf{m}_s
    \end{bmatrix},
\end{equation}
where $\gamma>0$ is the scaling constant, which can be fixed or trained.
Note that since geodesics in the Klein model are Euclidean lines and the isometry $\varphi_\mathbb{K}$ that maps from the Lorentz model to the Klein model is given by $\varphi_\mathbb{K}(\x) = {\x_s}/{x_t}$, this scaling simply slides $\mathbf{m}$ along the geodesic to or from the origin, ensuring the scaled output still satisfies \cref{main_prop}. Note that the scaling also levitates the potential limitation that the output of \method is always bounded by the larger of the two inputs, allowing for more expressiveness and representative power. 
\begin{figure}
   \label{many layers}
   \centering
   \begin{subfigure}[b]{0.4\textwidth}
\includegraphics[width=\linewidth]{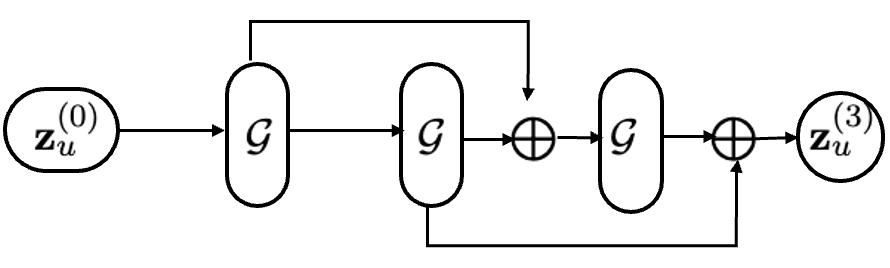}
    \caption{\method for GCN}\label{SkipGCN Visual}
    \label{Graph Results}
\end{subfigure}
    \centering
   \begin{subfigure}[b]{0.18\textwidth}
    \includegraphics[width=0.9\linewidth]{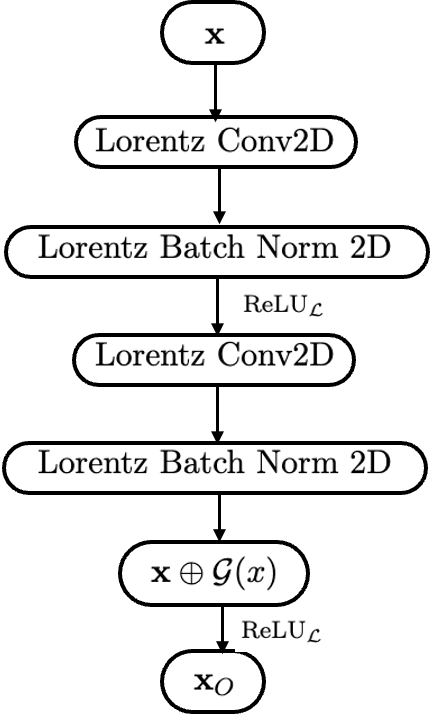}
    \caption{\method for CNN}\label{ResNet Visual}
\end{subfigure}
\centering
   \begin{subfigure}[b]{0.26\textwidth}
    \includegraphics[width=\linewidth]{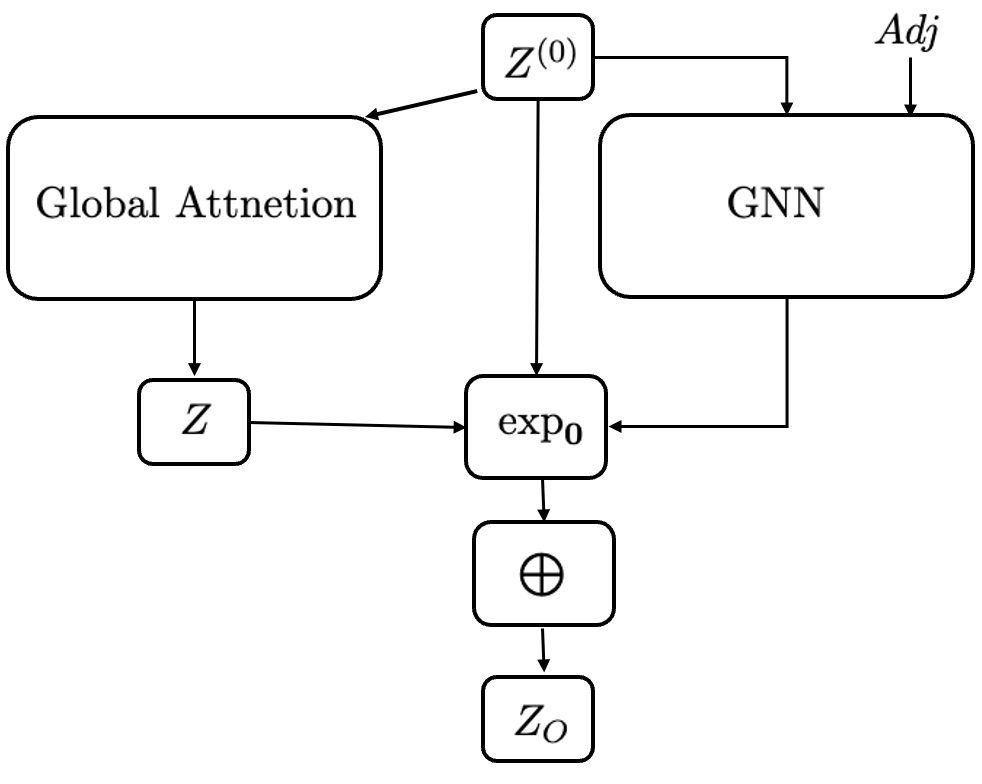}
    \caption{\method for Graph Transformer}\label{hyp gt}
    \end{subfigure}
\caption{Adaptation of \method to (a) 3-layer GNN architecture (b) residual block for vision tasks, and (c) graph transformer. $\mathcal{G}$ represents graph convolutional layers in (a) and convolutional layer implemented with HL in (b). $\bigoplus$ is a hyperbolic residual connection.}
\end{figure}
\section{Experiments} 
To demonstrate the effectiveness and versatility of our method, we apply \method as the residual connection in multiple adaptations of hyperbolic neural networks. We demonstrate that our method achieves better results than any previous methods of residual connection. In each task, we focus on testing the residual connection methods by using a consistent base model. In \cref{GNNs Exp} and \cref{trans}, we test \method on several graph datasets, as part of hyperbolic GNN and graph Transformer architectures to demonstrate its effectiveness. In \cref{Image Results}, we test \method on image datasets and demonstrate its effectiveness and robustness. In \cref{analysis}, we perform further analysis that demonstrates the efficiency and effectiveness of \method and shortcomings of previous methods. The datasets were selected as they exhibit low Gromovs $\delta$-hyperbolicity ($\delta$) \cite{hgcn2019} values, indicating highly hierarchical structures\footnote{For homogeneous graph datasets, $\delta$ is taken directly from HGCN~\cite{hgcn2019}. For heterogeneous datasets, $\delta$ was computed by us using the same code from HGCN~\cite{hgcn2019}. In both cases, the values were not normalized. For image datasets, $\delta$ is taken from HCNN\cite{Bdeir2024fully}, where the value is normalized by the largest distance between two points in the dataset using the method from~\cite{khrulkov2020hyperbolic}.}.

For all of our experiments, we used largely the same set of hyperparameters as we did the base model without residual connection whenever available. Additionally, we find that the addition of a residual layer does not add much tuning overhead. We include a discussion of the effects of the curvature $K$ as hyperparameters in \cref{{curv_hyp}}, which is missing in the work of the base models.

\subsection{Adaptation to GNNs } \label{GNNs Exp}
\textbf{GNN model architecture.} We formulate a skip-connected graph convolutional network with \method as the residual connection, using the fully hyperbolic graph convolutional layer~\cite{chen2021fully}. For the overall model architecture, we follow the ResGCN architecture outlined in~\cite{sun2021hgcf,yang2022hicf}, where the output of each intermediate layer was added to the output of the next layer via \method or a baseline hyperbolic residual connection method.
The model architecture is shown in \cref{SkipGCN Visual}, where $\mathcal{G}$ represents the graph convolutional layer and arcs are residual connections.

\begin{table*}
\caption{Test ROC AUC results (\%) for Link Prediction (LP),  F1 scores (\%) for Node Classification (NC) for homophilous graph, and accuracy (\%) for NC for heterophilic graph. The best performance is highlighted in bold. A lower 
$\delta$ value indicates a more tree-like dataset. The first four delta values are sourced from HGCN~\cite{hgcn2019}, and we used the same code for the remaining datasets.}
\label{graph table}
\centering
\resizebox{0.99\textwidth}{!}{%
\begin{tabular}{lcccccccccc}
\toprule
    & \multicolumn{8}{c}{\textbf{Homophilous}}   & \multicolumn{2}{c}{\textbf{Heterophilic}}            \\  
                   \textbf{Dataset}& \multicolumn{2}{c}{\sc{Disease}} & \multicolumn{2}{c}{\sc{Airport}} & \multicolumn{2}{c}{\sc{PubMed}} & \multicolumn{2}{c}{\sc{Cora}} & \sc{Chameleon}      & \sc{Squirrel} 
                   \\
                   \textbf{Hyperbolicity}& \multicolumn{2}{c}{$\delta=0$} & \multicolumn{2}{c}{$\delta=1$} & \multicolumn{2}{c}{$\delta=3.5$} & \multicolumn{2}{c}{$\delta=11$} & \sc{$\delta=2$}      & \sc{$\delta=1.5$}\\ 
\textbf{Task}      & LP                             & NC                               & LP                              & NC                              & LP                               & NC                              & LP                             & NC                             & NC                       & NC                       \\ \midrule
HyboNet~\cite{chen2021fully}           
& ${96.8 \pm 0.4}$                
& ${\mathbf{96.0} \pm 1.0}$                  
& ${\mathbf{97.3} \pm 0.3}$          
& ${90.0 \pm 1.4}$                 & ${95.8 \pm 0.2}$                  & ${78.0 \pm 1.0}$                 & ${93.6 \pm 0.3}$                & ${\mathbf{80.2} \pm 1.3}$                & ${40.1 \pm 0.8}$
& ${34.3 \pm 0.5}$  
\\
Parallel Transport~\cite{van2023poincar} & ${86.4 \pm 0.8}$                & ${84.8 \pm 3.7}$                  & ${93.6 \pm 0.1}$                 & ${93.4 \pm 0.6}$                 & ${\mathbf{96.5} \pm 0.1}$         & ${77.8 \pm 0.5}$                 & ${\mathbf{94.8} \pm 0.3}$       & ${76.0 \pm 0.8}$                  & ${36.6 \pm 1.9}$          & ${32.3 \pm 0.8}$          \\
Tangent Space~\cite{hgcn2019}      & ${76.0 \pm 2.4}$                  & ${91.9 \pm 1.9}$                  & ${93.5 \pm 0.1}$                 & ${92.0 \pm 2.9}$                   & ${96.4 \pm 0.2}$                  & ${76.8 \pm 0.9}$                 & ${94.1 \pm 0.3}$                & ${79.2 \pm 0.1}$                & ${38.3 \pm 0.8}$          & ${34.0 \pm 0.6}$          \\
Space Addition~\cite{Bdeir2024fully}     & ${83.1 \pm 1.2}$                & ${88.9 \pm 2.5}$                  & ${95.8 \pm 0.3}$                 & ${90.0 \pm 1.4}$                 & ${95.5 \pm 0.2}$                  & ${75.9 \pm 0.9}$                 & ${93.2 \pm 0.2}$                & ${78.6 \pm 0.5}$                & ${39.4 \pm 2.0}$          & ${34.5 \pm 0.2}$ 
\\
\textbf{\method (ours) }    & ${\mathbf{97.3} \pm 0.4}$          & ${\mathbf{96.1} \pm 1.0}$         & ${\mathbf{97.3} \pm 0.3}$        & ${\mathbf{93.9} \pm 0.7}$        & ${96.2 \pm 0.1}$                  & ${\mathbf{80.1} \pm 1.0}$        & ${94.1 \pm 0.3}$                & ${\mathbf{80.6} \pm 0.9}$       & ${\mathbf{41.1} \pm 0.9}$ & ${\mathbf{37.1} \pm 1.1}$ \\ \bottomrule
\end{tabular}%
}
\end{table*}

\textbf{GNN experimental setup. }
We evaluate \method on both node classification and link prediction tasks. We utilize several datasets: (1) homophilous graphs, including {\sc Disease}\cite{hgcn2019}, {\sc Airport}\cite{hgcn2019}, and two benchmark citation networks, namely {\sc PubMed}\cite{sen2008collective} and {\sc Cora}\cite{sen2008collective}; (2) Heterophilic graphs, including {\sc Chameleon} and {\sc Squirrel}, where we use the splits proposed by \cite{platonov2023critical}. For homophilous graphs, the {\sc Disease} and {\sc Airport} datasets exhibit a more hierarchical structure, whereas the citation networks are less so, making them suitable for demonstrating the generalization capability of \method. For evaluation, we report the F1-score for the node classification tasks and the Area Under Curve (AUC) for the link predictions task. For heterophilic graphs, we focus on the more difficult task of node classification and report accuracy percentage.

We closely follow the implementation of HyboNet model~\cite{chen2021fully}, utilizing its fully hyperbolic linear and aggregation layers. Our optimization follows the same setup, with parameters grouped based on their presence on the hyperbolic manifold. For the decoder, we implemented the Fermi-Dirac decoder~\cite{krioukov2009curvature, nickel2018learning}. For the implementation of \method, we applied constant weights of 1 for simplicity. 

\textbf{Baselines. }
We test the effectiveness of \method against other hyperbolic residual methods, by applying the baselines instead of \method. We consider the base HyboNet~\cite{chen2021fully} without residual connection, and the previous residual connection methods of the parallel transport method~\cite{hgcn2019}, the tangent space method~\cite{hgcn2019}, the space addition method~\cite{Bdeir2024fully}. 

\textbf{Experimental findings. }
We show the results in \cref{graph table}. Due to space constraints and for readability purposes, we omit the comparison to results from Euclidean GCNs and other hyperbolic GCNs. These comparisons can be found in Table 5 of~\cite{chen2021fully}, where the baseline HyboNet outperforms the aforementioned GCNs. As shown in \cref{graph table}, \method is the best performer in 8 out of the 10 tasks, for up to 4.2\% in the case of node prediction for {\sc Chameleon}. Compared to the base HyboNet without residual connection, \method substantially outperforms in 9 out of the 10 tasks, with the one remaining task being of comparable performance. Compared to the baseline residual connection methods, \method is the notably best performer in 8 out of the 10 tasks, especially for the more difficult tasks of node classification on heterophilic datasets, demonstrating its effectiveness and generalizability to more difficult problems. \method is also the best performer in every node prediction task, which benefits from deeper networks, demonstrating its superiority as a residual connection. In the more hyperbolic datasets, \method always performs better and by large margins, suggesting that it is more suitable for hyperbolic networks as it doesn't map between hyperbolic and tangent (Euclidean) spaces.  

\subsection{Adaptation to Graph Transformers}\label{trans}
We also investigate the application of our method to graph Transformers, where we test \method as part of a hyperbolic adaptation of SGFormer~\cite{wu2023sgformer}, a recent SOTA Euclidean graph Transformer. We consider the same hyperbolic residual connection baselines as did in \cref{GNNs Exp}, namely the parallel transport method, the tangent space method, and the space addition method.

\textbf{Lorentzian graph Transformer architecture. } Following the notations in SGFormer~\cite{wu2023sgformer}, let $\mathbf{Z}^{(0)}$ be the input embedding, $\mathbf{Z}$ be the output of the global attention layer, and $\mathrm{GN}(\mathbf{Z^{(0)}}, \mathbf{A})$ be the output of the GNN layer where $\mathbf{A}$ is the adjacency matrix.  For \method and the space addition method, we can project $\mathbf{Z}$ and $\mathrm{GN}(\mathbf{Z}^{(0)}, \mathbf{A})$ directly to $\mathbb{L}^{K,n}$. The final embedding is computed as \begin{equation}
    (1-\alpha)\exp_{\mathbf{o}}^K(\mathbf{Z}) \oplus \alpha\exp_{\mathbf{o}}^K(\mathrm{GN}(\mathbf{Z^{(0)}}, \mathbf{A})),
\end{equation}
where $\oplus$ denotes the respective residual connection method and $\alpha$ is a fixed weight. For the parallel transport and tangent space method, due to their dependence on the tangent space and the non-linearity of the exponential and logarithmic maps, we project weighted embeddings instead. In this case, the final embedding is then computed as \begin{equation}
    \exp_{\mathbf{o}}^K((1-\alpha)\mathbf{Z})) \oplus \exp_{\mathbf{o}}^K(\alpha\mathrm{GN}(\mathbf{Z^{(0)}}, \mathbf{A})).
\end{equation}
The final embedding of the model is then in hyperbolic space. \cref{hyp gt} shows the visualization of the model.

\textbf{Experimental setup.}
We closely followed the setup of the base SGFormer model~\cite{wu2023sgformer}. For the datasets, we consider 3 heterophilic graphs datasets, namely {\sc Chameleon}, {\sc Squirrel}, and {\sc Actor}. We use the same splits for {\sc Actor} as~\cite{lim2021new} and the same splits from earlier for {\sc Chameleon} and {\sc Squirrel}. We report node classification accuracy as did in SGFormer~\cite{wu2023sgformer}.

As for the decoder, we use the same Fermi-Dirac decoder~\cite{krioukov2009curvature, nickel2018learning} from earlier for the final fully connected layer (we elaborate on the detials for the decoder in the next paragraph). For the optimizer, we utilized two separate optimizers similar to HyboNet~\cite{chen2021fully} and the GCN experiments: the Euclidean Adam optimizer for non-manifold parameters and the Riemannian Adam optimizer for manifold parameters. We use trainable weights for \method. 

Here we elaborate on the decoder used for node classification for clarity. Specifically, let $\x_h$ be the final embedding preceding the decoder layer. Then the decoder layers learn $d$ vectors in hyperbolic space $\mathbf{v}_1,\ldots, \mathbf{v}_d$ to be the boundary of the $d$ node classes. Intuitively, the distance between $\x_h$ and $\mathbf{v}_k$ represents the negative log-likelihood of $\x_h$ being in class $k$. The classification of $\x_h$ is thus $\arg\max -d(\mathbf{v}_k, \x_h)$. Please see Section 3.3 of~\cite{nickel2018learning} for more details.

\begin{table}[tb]
\centering
\caption{Test accuracy {(\%)} for Node Classification on heterophilic graph datasets in a SGFormer-based graph Transformer architecture.} \label{transformer_results}
\resizebox{0.45\textwidth}{!}{
\begin{tabular}{@{}lcccc@{}}
\toprule
\textbf{Dataset}    & \sc{Chameleon}                     & \sc{Squirrel}                      & \sc{Actor}                                      &  \\ 
                   \textbf{Hyperbolicity} & $\delta = 2$                         & $\delta = 1.5$                         & $\delta = 1.5$                                      &  \\ \midrule
SGFormer~\cite{wu2023sgformer}           & $44.9 \pm 3.9$                           & $41.8 \pm 2.2$                           & $\mathbf{37.9} \pm 1.1$                                        &  \\
Parallel Transport~\cite{van2023poincar} & $46.7 \pm 1.2$                           & $38.5 \pm 1.3$                           & $35.5 \pm 0.9$                                        &  \\
Tangent Space~\cite{hgcn2019}      & $47.0 \pm 0.8$                           & $42.1 \pm 1.2$                           & $34.9 \pm 0.7$                                        &  \\
Space Addition~\cite{Bdeir2024fully}     & $47.2 \pm 1.4$                           & $43.0 \pm 1.1$                           & $35.3 \pm 0.4$                                        &  \\
\textbf{\method (ours)} & $\mathbf{47.8} \pm 1.3$ & $\mathbf{43.9} \pm 0.8$ & $\textbf{38.0}  \pm 0.4$ &  \\ \bottomrule
\end{tabular}}
\end{table}

 \begin{table}

\caption{Test accuracy (ACC) ($\%$) for Image Classification task on CIFAR-10 and CIFAR-100 in the hyperbolic ResNet. The $\delta$ values are taken from HCNN~\cite{Bdeir2024fully}, which are normalized by the diameter of the dataset.}
\label{Classification Results}
\centering
\resizebox{0.42\textwidth}{!}{%
\begin{tabular}{@{}lcc@{}}
\toprule
\textbf{Dataset}                            & \sc{CIFAR-10} (ACC) & \sc{CIFAR-100} (ACC) \\ 
\textbf{Hyperbolicity}                     &$\delta=0.26$ &$\delta=0.23$\\
\midrule
Parallel Transport~\cite{van2023poincar}         & $94.1\pm0.13$             & $72.9\pm 0.23$             \\ 
Tangent Space~\cite{hgcn2019}              & $94.0\pm0.19$             & $71.5\pm 0.30$             \\
Space Addition~\cite{Bdeir2024fully}            & $94.3\pm0.11$             & $74.3\pm0.22$              \\
\textbf{\method(ours)} & $\mathbf{94.6\pm0.17}$    & $\mathbf{74.8\pm0.25}$ \\ \bottomrule
\end{tabular}%
}
\end{table}

\begin{table*}[t]
\caption{ROC AUC (AUROC,\%), AUPR(\%), and FPR96(\%) results for OOD detection on CIFAR-10 and CIFAR-100 with Places365, DTD, and SVHN, as OOD datasets. R20 and R32 here denote ResNet-20 and ResNet-32 architectures, both with channel widths (8, 16, 32). The best performance is highlighted in bold. For FPR95, lower is better. For AUROC and AUPR, higher is better.}
\label{OOD Table}
\resizebox{0.70\textwidth}{!}{%
\begin{tabular}{@{}llllllllllllll@{}}
\toprule
\textbf{Dataset}               & \textbf{ResNet}               & \multicolumn{6}{c}{\textbf{CIFAR-10}}                                                                                                                                                                                                & \multicolumn{6}{c}{\textbf{CIFAR-100}} \\
               &               & \multicolumn{6}{c}{$\delta=0.26$}                                                                                                                                                                                                & \multicolumn{6}{c}{$\delta=0.23$} 
\\ \midrule
                               &                                 & \multicolumn{2}{c}{FPR95$\downarrow$}                                & \multicolumn{2}{c}{AUROC $\uparrow$}                                          & \multicolumn{2}{c}{AUPR $\uparrow$}                                          & \multicolumn{2}{c}{FPR95$\downarrow$}                                         & \multicolumn{2}{c}{AUROC $\uparrow$}                                 & \multicolumn{2}{c}{AUPR $\uparrow$}                                 \\
                               &                                 & R20                          & \multicolumn{1}{c}{R32}               & \multicolumn{1}{c}{R20}               & \multicolumn{1}{c}{R32}               & \multicolumn{1}{c}{R20}               & \multicolumn{1}{c}{R32}               & R20                                   & \multicolumn{1}{c}{R32}               & \multicolumn{1}{c}{R20}      & \multicolumn{1}{c}{R32}               & \multicolumn{1}{c}{R20}      & \multicolumn{1}{c}{R32}               \\ \midrule
{Places} & Euclidean \cite{he2016deep}                       & $64.2$                         & $72.3$                                  & $\mathbf{84.7}$                         & $82.0$                                  & $\mathbf{96.2}$                         & $95.6$                                  & $89.5$                                  & $93.9$                                  & $62.5$                         & $57.9$                                  & $89.3$                         & $87.9$                                  \\
                               & w/ HNN++\cite{van2023poincar}                        & $63.8$                & $72.7$                                  & $79.6$                                  & $77.7$                                  & $94.5$                                  & $94.2$                                  & $93.2$                                  & $86.3$                                  & $63.6$                         & $66.6$                                  & $89.8$                         & $91.1$                                  \\
                               & Poincar{\'e}\cite{van2023poincar}                         & $70.2$                         & $70.7$                                  & $82.3$                                  & $82.6$                                  & $95.7$                                  & $95.9$                         & $\mathbf{82.8}$                         & $\mathbf{83.8}$                         & $\mathbf{71.5}$                & $71.1$                        & $\mathbf{92.3}$              & $92.2$                                  \\ 
                               & \cellcolor[HTML]{E8E8E8}\textbf{\method(ours)} & \cellcolor[HTML]{E8E8E8}$\mathbf{63.0}$ & \cellcolor[HTML]{E8E8E8}$\mathbf{63.3}$ & \cellcolor[HTML]{E8E8E8}$82.6$          & \cellcolor[HTML]{E8E8E8}$\mathbf{82.7}$ & \cellcolor[HTML]{E8E8E8}$95.7$          & \cellcolor[HTML]{E8E8E8}$\mathbf{96.0}$ & \cellcolor[HTML]{E8E8E8}$93.2$          & \cellcolor[HTML]{E8E8E8}$93.2$          & \cellcolor[HTML]{E8E8E8}$53.6$ & \cellcolor[HTML]{E8E8E8}$\mathbf{71.7}$         & \cellcolor[HTML]{E8E8E8}$84.5$ & \cellcolor[HTML]{E8E8E8}$\mathbf{92.4}$ \\ \midrule
SVHN                           & Euclidean\cite{he2016deep}                      & $97.3$                         & $94.7$                                  & $68.8$                                  & $73.4$                                  & $92.8$                                  & $94.1$                                  & $99.5$                                  & $98.8$                                  & $43.7$                         & $54.6$                                  & $83.7$                         & $88.2$                                  \\
                               & w/ HNN++\cite{van2023poincar}                         & $73.1$                         & $79.1$                                  & $85.5$                         & $82.2$                                  & $96.9$                         & $96.1$                                  & $92.1$                                  & $88.6$                                  & $66.4$                         & $68.9$                                  & $91.1$                         & $92.0$                                  \\
                               & Poincar{\'e}\cite{van2023poincar}                         & $66.0$                & $69.3$                                  & $85.0$                                  & $83.6$                                  & $96.6$                                  & $96.3$                                  & $\mathbf{76.9}$                         & $83.0$                                  & $\mathbf{76.8}$                & $72.6$                         & $\mathbf{94.1}$                & $\mathbf{92.9}$                         \\
                               & \cellcolor[HTML]{E8E8E8}$\mathbf{\method(ours)}$ & \cellcolor[HTML]{E8E8E8}$\mathbf{39.0}$ & \cellcolor[HTML]{E8E8E8}$\mathbf{56.9}$ & \cellcolor[HTML]{E8E8E8}$\mathbf{90.8}$         & \cellcolor[HTML]{E8E8E8}$\mathbf{88.0}$ & \cellcolor[HTML]{E8E8E8}$\mathbf{98.0}$          & \cellcolor[HTML]{E8E8E8}$\mathbf{97.4}$ & \cellcolor[HTML]{E8E8E8}$80.1$          & \cellcolor[HTML]{E8E8E8}$\mathbf{80.7}$ & \cellcolor[HTML]{E8E8E8}$66.9$ & \cellcolor[HTML]{E8E8E8}$\mathbf{73.4}$          & \cellcolor[HTML]{E8E8E8}$90.7$ & \cellcolor[HTML]{E8E8E8}$\mathbf{92.9}$ \\ \midrule
Texture                        & Euclidean\cite{he2016deep}                       & $87.3$                         & $88.0$                                  & $73.6$                                  & $77.3$                                  & $93.2$                                  & $94.7$                                  & $98.1$                                  & $96.0$                                  & $33.5$                         & $42.9$                                  & $75.9$                         & $79.4$                                  \\
                               & w/ HNN++\cite{van2023poincar}                         & $\mathbf{63.8}$                & $\mathbf{56.6}$                         & $79.6$                                  & $\mathbf{85.8}$                         & $94.5$                                  & $96.6$                                  & $85.9$                                  & $\mathbf{77.5}$                        & $58.9$                         & $65.7$                                  & $86.8$                         & $89.0$                                  \\
                               & Poincar{\'e}\cite{van2023poincar}                        & $68.2$                         & $66.2$                                  & $82.1$                                  & $82.3$                                  & $95.5$                                  & $95.6$                                  & $83.9$                                  & $84.2$                                  & $\mathbf{67.7}$                & $68.8$                                  & $\mathbf{91.0}$                & $\mathbf{91.5}$                         \\
                               & \cellcolor[HTML]{E8E8E8}\textbf{\method(ours)} & \cellcolor[HTML]{E8E8E8}$\mathbf{63.8}$ & \cellcolor[HTML]{E8E8E8}$68.4$          & \cellcolor[HTML]{E8E8E8}$\mathbf{84.2}$ & \cellcolor[HTML]{E8E8E8}$\mathbf{85.8}$          & \cellcolor[HTML]{E8E8E8}$\mathbf{96.3}$ & \cellcolor[HTML]{E8E8E8}$\mathbf{96.7}$ & \cellcolor[HTML]{E8E8E8}$\mathbf{83.7}$ & \cellcolor[HTML]{E8E8E8}$83.7$          & \cellcolor[HTML]{E8E8E8}$57.6$ & \cellcolor[HTML]{E8E8E8}$\mathbf{69.1}$ & \cellcolor[HTML]{E8E8E8}$86.3$ & \cellcolor[HTML]{E8E8E8}$\mathbf{91.5}$ \\ \bottomrule
\end{tabular}%
}
\end{table*}

\textbf{Experimental findings. } The results are presented in \cref{transformer_results}. \method consistently outperforms both the base Euclidean SGFormer and the baseline hyperbolic residual connection methods across all three cases, highlighting its effectiveness in Transformer models. Moreover, in 2 of the 3 datasets, the hyperbolic SGFormer nearly always surpasses the Euclidean version, supporting the advantages of hyperbolic modifications.

\subsection{Adaptation to Vision Model} \label{Image Results}

 \textbf{Lorentzian ResNet architecture. }
Here we define the hyperbolic ResNet structure we use for the image experiments. The network is made simple to highlight the effects of residual connections. For the traditional convolutional layer, batch normalization layer, and ReLU activation layer utilized in Euclidean ResNet in~\cite{he2016deep}, we consider a simple hyperbolic adaptation. For an input vector $\x\in\mathbb{L}^{K,n}$, we define a \textit{hyperbolic layer} (HL) to be a neural network layer that first applies the Euclidean layer on the space-like dimension of an input vector $\x$ to obtain $f(\x_s)$ where $f$ is the Euclidean layer, and then calculating the time-like dimension afterward. Formally, it outputs a vector $\y\in\mathbb{L}^{K,n}$ such that $\y_s = f(\x_s)$ and $ y_t = \sqrt{||\y_s||^2 - \frac{1}{K}}$. For \method, we use trainable weights as outlined in \cref{midpoint formula}. We also use the scaling method outlined in \cref{scaling formula} to take advantage of the positive correlation between embedding norm and classification confidence~\cite{ghadimi2021hyperbolic}. \cref{ResNet Visual} shows a visualization of our residual block, where the Lorentz variation (and subscript $\mathcal{L}$) of an operation indicates implementation with a hyperbolic layer. For the decoder, we experienced numerical instability with the Lorentz MLR in HCNN\cite{Bdeir2024fully}. As a result, we use the isometry map \cite{hgcn2019} to project the final embedding from the Lorentz model into the Poincar{\'e} ball model to use the Poincar{\'e} MLR from HNN++\cite{HNN++}. Specifically, for the final fully connected output layer, we first map the Lorentzian embedding to the corresponding Poincar{\'e} ball model, then we use the pooling method in Poincar{\'e} ResNeT~\cite{van2023poincar} where pooling is done in Euclidean space and mapped back into the Poincar{\'e} ball. The Poincar{\'e} MLR layer is then applied to the hyperbolic pooling output. For the optimizer, we used the Riemannian SGD optimizer that was used in HCNN~\cite{Bdeir2024fully} for classification and the Adam optimizer for ODD-detection as did in Poincar{\'e} ResNet~\cite{van2023poincar}. 

\textbf{Classification accuracy experiment. }
To demonstrate the effectiveness of \method over previous methods, we evaluate image classification performance using the ResNet-18 architecture \cite{he2016deep} and two datasets: CIFAR-10\cite{Krizhevsky2009Learning} and CIFAR-100\cite{Krizhevsky2009Learning}. For the experiment, we use the structure shown in \cref{ResNet Visual} and test our residual connection against previous methods. For the baseline, we test against the parallel transport method, the tangent space method, and the space addition method. As parallel transport is not commutative, we use $\x\oplus_P f(\x)$ where $f(\x)$ is the output of the convolutional layer. We adopt the training procedure of \cite{Bdeir2024fully, DeVries2017improved}
which have been optimized for Euclidean CNNs and yielded strong results for both Euclidean and Hyperbolic ResNets. The results are shown in \cref{Classification Results}. \method performs notably better than all of the previous methods for both datasets, demonstrating its effectiveness as a residual connection method for vision tasks. 

\textbf{Out-of-distribution (OOD) robustness. }
To check that \method is robust to out-of-distribution (OOD) samples, we test the OOD detection performance of our Lorentzian ResNet-20 and ResNet-32 with channel widths (8, 16, 32), trained on either CIFAR-10 or CIFAR-100. We closely follow the experimental setup in Poincar\'e ResNet~\cite{van2023poincar}. We use the Places-365 dataset \cite{zhou2017places}, the Texture dataset \cite{cimpoi2014texture}, and the SVHN dataset \cite{netzer2011reading}. The baseline we compare to are the Euclidean and hyperbolic ResNets in Poincar\'e ResNet~\cite{van2023poincar}. Poincar\'e ResNet is a baseline hyperbolic ResNet model using specially designed hyperbolic convolutional layers and uses the parallel transport method as residual connection, thus it can be seen as a representative of the parallel transport baseline. As for the detection of the OOD sample, we use the energy score which was introduced by~\cite{liu2020energy} and also used in~\cite{van2023poincar}. Finally, we compare our results against the baselines with the three commonly used metrics of FPR95, AUROC, and AUPR. 

\textbf{OOD detection results. }
The results in \cref{OOD Table} indicate that \method, when trained on CIFAR-10, significantly outperforms Euclidean counterparts and previous hyperbolic ResNets on almost all metrics. On CIFAR-100, our model notably surpasses the baseline with a ResNet-32 architecture. This demonstrates that \method is more robust to OOD samples than its Euclidean and hyperbolic counterparts using parallel transport for residual connections, confirming that \method is not only effective and reliable.

\begin{table}[tb]
\centering
\caption{Test ROC AUC result (\%) for both directions of the parallel transport (PT) method in Link Prediction.}
\label{non commute table}
\resizebox{0.4\textwidth}{!}{
\begin{tabular}{@{}lllll@{}}
\toprule
\textbf{Direction of PT } & \multicolumn{2}{c}{Disease ($\delta=0$)} & \multicolumn{2}{c}{Airport ($\delta=1$)} \\
\midrule
Forward           & \multicolumn{2}{c}{$\mathbf{86.8 \pm 0.8}$} & \multicolumn{2}{c}{$93.6 \pm 0.1$}       \\
Backward          & \multicolumn{2}{c}{$67.7 \pm 1.4$}          & \multicolumn{2}{c}{$\mathbf{93.7 \pm 0.1}$} \\ \bottomrule
\end{tabular}
}
\end{table}

\subsection{Further Analysis} 
\label{analysis}
This section presents a detailed analysis of our proposed method, $\method$, highlighting its high efficiency through runtime experiments. We also examine its effectiveness in addressing the over-smoothing issue in deeper graph neural networks. Additionally, We demonstrate the need for a commutative residual connection by examining the effects of non-commutativity on the flexibity and expressiveness of the parallel transport method.
\begin{table}[tb]
\caption{Average runtime to perform 100 additions on random vectors of dimension 2,048 and dimension size 10,000, and dimension 4,096 and size 100,000, on an RTX 3070.}\label{time table}
\centering
\resizebox{0.4\textwidth}{!}{
\begin{tabular}{@{}lcccc@{}}
\toprule
\textbf{Method}    & $2,048/10,000$ & $4,096/100,000$ \\ \midrule
Parallel Transport~\cite{van2023poincar} & $0.0036$s     & $1.448$s       \\
Tangent Space~\cite{hgcn2019}       & $0.0083$s     & $3.601$s       \\
\textbf{LResNet (ours) } & $0.00025$s    & $0.0006$s      \\
\bottomrule
\end{tabular}\label{time_analysis}
}
\end{table}

\textbf{Efficiency analysis. }
We compare the efficiency of \method with previous multi-mapping methods, specifically the parallel transport and tangent space method, by conducting 100 additions on randomly generated hyperbolic vectors (on the manifold with curvature $K = -1$) with dimensions of 2048 and sizes of 10,000, and dimensions of 4096 and sizes of 100,000, using a single RTX 3070 GPU. The average runtime for each method is presented in \cref{time_analysis}. \method demonstrates significant speedup over the baselines and offers better scalability as the size and dimension increase, for over $2000$ times speedup. 

\begin{figure}
   \centering
   \begin{subfigure}[b]{0.22\textwidth}
       \includegraphics[width=\linewidth]{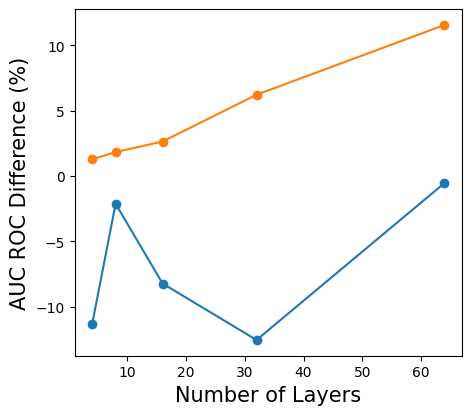}
       \caption{Disease}
       \label{fig:disease_many_layers}
   \end{subfigure}
    \centering
   \begin{subfigure}[b]{0.22\textwidth}
       \includegraphics[width=\linewidth]{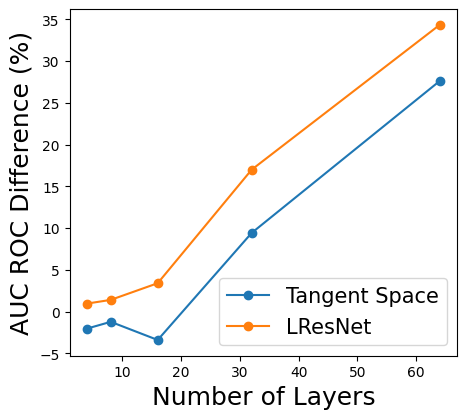}
       \caption{Airport}
       \label{fig:airport_many_layers}
   \end{subfigure}
   \caption{Comparison of ROC AUC (\%) differences for link prediction (LP) between HyboNet with and without residual connections: orange indicates our \method as the residual connection, while blue represents the tangent space method as the residual connection. In \cref{fig:disease_many_layers} we show the results on the {\sc{Disease}} dataset and in \cref{fig:airport_many_layers} we show the results on the {\sc{airport}} dataset.}\label{many layers}
\end{figure}

\textbf{Overcoming graph over-smoothing. }
Previous studies have shown that stacking graph convolutional layers often leads to performance drops due to gradient vanishing or over-smoothing \cite{zhao2019pairnorm}. Residual connections traditionally address these issues in Euclidean space. To examine the impact of the proposed method in hyperbolic space, we implemented the GCN architecture from \cref{GNNs Exp} on the {\sc Disease} and {\sc Airport} datasets across 4, 8, 16, 32, and 64 layers, using the tangent space method as a baseline for link prediction tasks. The results in \cref{many layers} depict the difference in ROC AUC scores between residual-connected HyboNets and the base HyboNet for visual clarity. We observed that the performance gap between HyboNet and \method widens with more layers, indicating the effectiveness of \method in countering over-smoothing. Conversely, while the gap widens for the tangent space method as well, it shows a reduction in differences, especially at 32 layers for {\sc Disease} and 16 for {\sc Airport}, highlighting \method's stability when applied to deep neural networks. We found that in practice, parallel transport method yielded NaN values when using 16 or more layers, reflecting its numerical instability. The results were therefore excluded.

\textbf{Commutative capability.} \label{non-commute}
To investigate the effect of commutativity of the residual connection in hyperbolic models, we test the performance on link prediction of both directions of addition for the parallel transport method on {\sc Disease} and {\sc Airport} datasets using the architecture in \cref{SkipGCN Visual}. Forward operation represents $\x\oplus_P f(\x)$ and backward operation represents $f(\x)\oplus_P \x$, where $f(\x)$ is the output of the hyperbolic graph convolutional layer. The results are shown in \cref{non commute table}. The forward method significantly outperforms the backward method in link prediction tasks of the {\sc Disease} dataset, while the backward method outperforms the forward method for the {\sc Airport} dataset. This shows the limitation in the flexibility of the parallel transport method due to its non-commutativity, as it is unpredictable which direction of addition would have the better performance.

\section{Conclusion}
In this work, we proposed \method, a hyperbolic residual neural network based on the weighted Lorentzian centroid in the Lorentz model of hyperbolic geometry. \method addresses the limitations of previous methods, offering increased efficiency, flexibility, numerical stability, commutativity, and geometric information retention. We proved that \method can theoretically derive previous methods while ensuring numerical stability.
Empirically, \method outperforms previous methods in image classification, link prediction, and node classification tasks, demonstrating superior performance and robustness compared to Euclidean and hyperbolic counterparts across multiple architectures, including GNNs, CNNs, and graph Transformers. However, it should also be noted that the performance of a residual-connected model is still conditioned upon the performance of the base model, while there lacks an abundance of hyperbolic models, such as in the case of diffusion models. One direction of future work is to apply \method to newly developed hyperbolic model architectures.

\begin{acks}
This work was supported in part by the National Science Foundation (NSF) IIS Div Of Information \& Intelligent Systems 2403317. We also gratefully acknowledge support in part from Amazon Research Award, Snap Research, and the Yale Office of the Provost. Additionally, this research has greatly benefited from the discussions and research talks held at the IMS-NTU Joint Workshop on Applied Geometry for Data Sciences.  We extend our sincere gratitude to the reviewers for their valuable feedback and insightful suggestions.
\end{acks}

\newpage
\bibliographystyle{ACM-Reference-Format}
\bibliography{main}
\appendix
\section{Proof of Theoretical Results}\label{proofs}
\subsection*{Proof of \cref{stability lemma}}
\begin{proof}
    Note that \begin{align*}
        &-K||w_x\x + w_y\y||_\mathcal{L}^2\\ &= -K\left(-(w_xx_t + w_yy_t)^2 + ||w_x\x_s + w_y\y_s||^2 \right)\\
        &= K||w_x\x||_\mathcal{L}^2 + K||w_y\y||_\mathcal{L}^2 - 2K\langle w_x\x, w_y\y\rangle_\mathcal{L}\\
        &= w_x^2 + w_y^2 - 2K\Gamma\\
        &> w_x^2 + w_y^2 - 2K(w_xw_y||\x_s||||\y_s|| - w_xw_y\langle\x_s.\y_s\rangle)\\
        &\geq w_x^2 + w_y^2,\tag{Cauchy Schwarz Inequality}
    \end{align*}
    where \begin{equation}
        \begin{split}
            \Gamma &= w_x\sqrt{\|\x_s\|^2 - \frac{1}{K}}\cdot w_y\sqrt{||\y_s||^2 - \frac{1}{K}} - w_xw_x\langle\x_s, \y_s\rangle
        \end{split}
    \end{equation}
    By taking the square roots we have the desired inequality.
\end{proof}
\subsubsection*{Proof of \cref{theorem:non-commutative}}
\begin{proof}
   Let $\x = [a_1,\ldots,-a_{n+1}]\in \mathbb{L}_K^n$ and $\y = [a_1,\ldots,a_{n+1}] \in \mathbb{L}_K^n$ where each $a_i\in\R$ and $a_1>0$. We can easily compute (by following the computation in \cref{main_prop}) that we have $$\z = \x \oplus\y = \cosh(\alpha)\x + \frac{\sinh(\alpha)}{\alpha}(c_u\y' + c_v\x'),$$ where 
\begin{align*}
    &\y' = \y +y_t\sqrt{-K}\mathbf{o}, \x' = \x + \mathbf{o},\\
    &c_u = \frac{\cosh^{-1}(y_t\sqrt{-K})}{\sqrt{-y_t^2K-1}}, \mathbf{u} = c_u\cdot\y'\\
    &c_v = \frac{\langle \x,\mathbf{u}\rangle_{\mathcal{L}}}{-1/K - \langle \mathbf{o}, \x\rangle_{\mathcal{L}}}, \mathbf{v} = c_u\cdot\y'+c_v\cdot\x'\\
    &\alpha = \sqrt{-K}||\mathbf{v}||_{\mathcal{L}}.
\end{align*}
Similarly, we can compute that for the other direction of parallel transport, we have $$\z' = \y\oplus \x = \cosh(\alpha')\y + \frac{\sinh(\alpha')}{\alpha'}(c_u'\mathbf{p} + c_v'\mathbf{q}),$$ where
\begin{align*}
    &\mathbf{p} = \x +x_t\sqrt{-K}\mathbf{o}, \mathbf{q} = \y + \mathbf{o}\\
    &c_u' = \frac{\cosh^{-1}(x_t\sqrt{-K})}{\sqrt{-x_t^2K-1}}, \mathbf{u}' = c_u'\cdot\mathbf{p}\\
    &c_v' = \frac{\langle \y,\mathbf{u}'\rangle_{\mathcal{L}}}{-1/K - \langle \mathbf{o}, \y\rangle_{\mathcal{L}}}, \mathbf{v}' = c_u'\cdot\mathbf{p}+c_v'\cdot\mathbf{q}\\
    &\alpha' = \sqrt{-K}||\mathbf{v}'||_{\mathcal{L}}.
\end{align*}
By analyzing the symmetry in the construction of $\mathbf{x}, \mathbf{y}, \mathbf{y}', \mathbf{p}, \mathbf{u}, \mathbf{u}', \mathbf{v},$ and $\mathbf{v}'$, it can be shown that $c_u = c_u'$ and $c_v = c_v'$ due to the identical nature of the transformations applied in either direction. This implies that $\alpha = \alpha'$, leading to the conclusion that $\mathbf{z}_{n+1} = -\mathbf{z}'_{n+1}$ as the operations in the parallel transport in both directions result in vectors whose $(n+1)$-th components are negations of each other. In the following, we give the detailed derivation.
 
First, since $x_t = y_t$, we have $c_u = c_u'$. Let $c_u = c_u'=\beta$, then we have,
\begin{align*}
    \y' = \y + y_t\sqrt{-K}\mathbf{o} &= \y + a_1\cdot\sqrt{-K}\left[\sqrt{-1/K},0,\ldots,0\right]^T\\
    &= [2a_1,a_2,\ldots,a_{n+1}]^T
\end{align*}
and
\begin{align*}
    \mathbf{p} = \x + x_t\sqrt{-K}\mathbf{o} &= \x + a_1\cdot\sqrt{-K}\left[\sqrt{-1/K},0,\ldots,0\right]^T\\
    &= [2a_1,a_2,\ldots,-a_{n+1}]^T.
\end{align*}
So $\mathbf{u} = \beta\y' = [2\beta a_1,\beta a_2,\ldots,\beta a_{n+1}]^T$ and $\mathbf{u}' = \beta\mathbf{p} = [2\beta a_1,\beta a_2, \ldots, -\beta a_{n+1}]$. So we can compute that
\begin{align*}
    &\langle \x, \mathbf{u}\rangle_{\mathcal{L}} = \left[-2\beta a_1^2 - \beta a_{n+1}^2 + \displaystyle\sum_{i = 2}^{n}\beta a_i^2\right]^T\\
    &\langle \y, \mathbf{u}'\rangle_{\mathcal{L}} = \left[-2\beta a_1^2 - \beta a_{n+1}^2 + \displaystyle\sum_{i = 2}^{n}\beta a_i^2\right]^T.
\end{align*}
So we have 
\begin{align*}
    c_v &= \frac{\langle \x,\mathbf{u}\rangle_{\mathcal{L}}}{-1/K - \langle \mathbf{o}, \x\rangle_{\mathcal{L}}}\\
   &= \frac{\left[-2\beta a_1^2 - \beta a_{n+1}^2 + \displaystyle\sum_{i = 2}^{n}\beta a_i^2\right]^T}{-1/K - \sqrt{-1/K}a_1}
\end{align*}
and
\begin{align*}
    c_v' &= \frac{\langle \y,\mathbf{u}'\rangle_{\mathcal{L}}}{-1/K - \langle \mathbf{o}, \y\rangle_{\mathcal{L}}}\\
    &= \frac{\left[-2\beta a_1^2 - \beta a_{n+1}^2 + \displaystyle\sum_{i = 2}^{n}\beta a_i^2\right]^T}{-1/K - \sqrt{-1/K}a_1}.
\end{align*}
So we have $c_v = c_v'$, call this constant $\gamma$. 
Now, we can compute further that 
\begin{align*}
    \mathbf{v} &= \beta \y' + \gamma\x' \\&= \beta[2 a_1,  a_2, \ldots,  a_{n+1}]^T \\&+ \gamma\left[(a_1 + \sqrt{-1/K}),a_2, \ldots, a_{n+1}\right]^T\\
    &= \left[2\beta a_1 + \gamma(a_1 + \sqrt{-1/K}),\ldots,(\beta + \gamma)a_{n+1}\right]^T.
\end{align*}
and
\begin{align*}
    \mathbf{v}' &= \beta \mathbf{p} + \gamma\mathbf{q} \\&= [2\beta a_1, \beta a_2, \ldots, -\beta a_{n+1}]^T \\&+ \left[\gamma(a_1 + \sqrt{-1/K}), \gamma a_2,\ldots, -\gamma a_{n+1}\right]^T\\
    &= \left[2\beta a_1 + \gamma(a_1 + \sqrt{-1/K}),\ldots,-(\beta + \gamma)a_{n+1}\right]^T.
\end{align*}
So we have $||\mathbf{v}||_{\mathcal{L}} = ||\mathbf{v}'||_{\mathcal{L}}$, and thus $\alpha = \alpha'$, let us call this constant simply $\alpha$. Finally:
\begin{align*}
    &\z_{n+1} = \cosh(\alpha)(-a_{n+1}) + \frac{\sinh(\alpha)}{\alpha}(\beta a_{n+1} -\delta a_{n+1})\\
    &\z'_{n+1} = \cosh(\alpha)(a_{n+1}) + \frac{\sinh(\alpha)}{\alpha}(\beta(-a_{n+1}) +\delta a_{n+1}).
\end{align*}
Hence, $\z_{n+1} = -\z'_{n+1}$.
\end{proof}

\subsubsection*{Proof of Proposition~\ref{main_prop}}
\begin{proof}
\begin{enumerate}[label=\alph*.]
        \item For \textbf{parallel transport method}:
     First note that the isometry from Lorentz model to Klein model is given by the map $$\varphi_K(\x) = \x_t/x_s.$$ Thus given the fact that geodesics in the Klein model are Euclidean straight lines, it suffices to show that $\mathbf{z}_s = \lambda\mathbf{m}_s$ for some $\lambda\in\mathbb{R}$. Then we can compute:
    \begin{align*}
        \mathbf{u}= \log_{\mathbf{o}}^K(\mathbf{y})&= \frac{\cosh^{-1}\left(y_t\sqrt{-K}\right)}{\sqrt{-y_t^2K-1}}(\mathbf{y}+y_t\sqrt{-K}\mathbf{o})\\ &= c_u\cdot\mathbf{y}'
    \end{align*}
    \begin{align*}
        \mathbf{v} = \mathbf{P}_{\mathbf{o}\to\mathbf{x}}(\mathbf{u}) &= \mathbf{u} + \frac{\langle\mathbf{x},\mathbf{u}\rangle_{\mathcal{L}}}{-1/K - \langle\mathbf{o},\mathbf{x}\rangle_{\mathcal{L}}}(\mathbf{o}+\mathbf{x}) \\&= c_u\cdot\mathbf{y}' + c_v\cdot \mathbf{x}'
    \end{align*}
    \begin{align*}
        \z &= \cosh(\sqrt{-K}||\mathbf{v}||_{\mathcal{L}})\mathbf{x} + \frac{\sinh(\sqrt{-K}||\mathbf{v}||_{\mathcal{L}})}{\sqrt{-K}||\mathbf{v}||_{\mathcal{L}}}\mathbf{v} \\&= \cosh(\alpha)\mathbf{x} + \frac{\sinh(\alpha)}{\alpha}(c_u\mathbf{y}' + c_v\mathbf{x}')
    \end{align*}
    where \[c_u = \frac{\cosh^{-1}\left(y_t\sqrt{-K}\right)}{\sqrt{-y_t^2K-1}},c_v = \frac{\langle\mathbf{x},\mathbf{u}\rangle_{\mathcal{L}}}{-1/K - \langle\mathbf{o},\mathbf{x}\rangle_{\mathcal{L}}},\] \[\mathbf{y}' = \mathbf{y} + y_t\sqrt{-K}\mathbf{o}, \mathbf{x}' = \mathbf{o} + \mathbf{x}, \alpha = \sqrt{-K}||\mathbf{v}||_\mathcal{L}.\] Clearly $\mathbf{y}'_s = \mathbf{y}_s, \mathbf{x}'_s = \mathbf{x}_s$. So $w_x = \cosh(\alpha) + c_v$ and $w_y = \frac{\sinh(\alpha)}{\alpha}c_u$ gives the desired result.
\item For \textbf{tangent space method}: One can check that \begin{align*}&\exp_{\mathbf{o}}^K(\log_{\mathbf{o}}^K(\mathbf{x}) + \log_{\mathbf{o}}^K(\mathbf{y}))_s \\&= \frac{\sinh(\sqrt{-K}||c_1\mathbf{x}' + c_2\mathbf{y}'||_\mathcal{L})}{\sqrt{-K}||c_1\mathbf{x}' + c_2\mathbf{y}'||_\mathcal{L}}\left(c_1\mathbf{x}' + c_2\mathbf{y}'\right)_s
\end{align*}
with $c_1 = \frac{\cosh^{-1}\left(-x_t\sqrt{-K}\right)}{\sqrt{x_t^2K-1}}, c_2 = \frac{\cosh^{-1}\left(-y_t\sqrt{-K}\right)}{\sqrt{y_t^2K-1}}, \mathbf{x}' = \mathbf{x} + x_t\sqrt{-K}\mathbf{o}, \mathbf{y}' = \mathbf{y} + y_t\sqrt{-K}\mathbf{o}$. Again $\x_s = \x'_s, \y_x = \y'_s$ So one can pick $w_x = c_1, w_y = c_2$.
\item For \textbf{space addition method}:
This follows immediately from the isometry to the Klein model.
\end{enumerate}
\end{proof}
\section{Dataset Processing Details}\label{datasets}
\subsubsection*{Graph datasets}
For the homophilous datasets, we used the same splits as in HGCN~\cite{hgcn2019}. For the heterophilic datasets, we use the same splits as in SFGormer~\cite{wu2023sgformer}, where the splits for {\sc Chameleon} and {\sc Squirrel} are from~\cite{platonov2023critical}, and the splits for {\sc Actor} are from~\cite{lim2021new}. 

\subsubsection*{Image datasets}
For the image classification tasks, we used the splits implemented in PyTorch, where there are 50,000 training images and 10,000 testing images. For OOD-dection, we followed the set up in~\cite{van2023poincar}, 
where we use the same data splits as in~\cite{van2023poincar} and~\cite{liu2020energy}.

\section{Implementation and Training Details} \label{training}

\begin{table}[]
\caption{AUC-ROC for LP and NC for varying curvature.}\label{many_curv}
\resizebox{0.4\textwidth}{!}{%
\begin{tabular}{lllll}
\hline
\textbf{Curvature} & \multicolumn{2}{c}{\sc{disease}} & \multicolumn{2}{c}{\sc{airport}} \\ \hline
                   & LP                         & NC                       & LP                        & NC                        \\
$-0.5$             & $96.7\pm0.4$               & $95.4\pm1.2$             & $96.8\pm0.2$              & $93.9\pm0.7$              \\
$-1.0$             & $97.1\pm0.3$               & $96.0\pm0.8$             & $97.0\pm0.3$              & $92.9\pm0.5$              \\
$-1.5$             & $97.1\pm0.3$               & $95.8\pm1.2$             & $97.3\pm0.3$              & $93.6\pm1.0$              \\
$-2.0$             & $97.3\pm0.4$               & $95.6\pm0.9$             & $97.3\pm0.1$              & $93.3\pm 0.8$             \\
Trainable          & $96.5\pm0.5$              & $96.1\pm1.0$             & $96.4\pm0.5$              & $93.1\pm1.0$              \\ \hline
\end{tabular}%
}
\end{table}
\subsection{Implementation details for graph datasets}
\subsubsection{GNN hyperparameters}
For the homophilous graphs datasets, we used the largely the same hyperparmeters as in~\cite{chen2021fully}, including dropout rate, learning rate, weight decay rate, gradient clip value, and margins for the marginal loss. For link prediction tasks, we run the model for 3 layers. For node classification tasks, we performed a search for the number of layers on the set $\{3, 4, 5, 6, 7, 8\}$. For all tasks, we performed a search for the constant negative curvature $K$ on the set $\{-0.1, -0.5, -1.0, -1.5, -2.0, \mathrm{trainable}\}$, where the initial value of the trainable curvature is $-1.0$. 

For heteophilic graph datasets, we used a curvature of $-1$ and performed a grid search in the following search space: dimension within $\{16, 32, 64\}$; learning rate within $\{0.001, 0.01\}$;dropout rate within $\{0, 0.1, 0.2, 0.3\}$; weight-decay rate within $\{0, 1e-4, 1e-3\}$; number of layers within $\{3, 4, 5, 6, 7, 8\}$. While many values were searched, we find that the performance of \method and the baselines depend mostly on the that of the base HyboNet without residual connection. As a result, we used the same hyperparameters as in~\cite{chen2021fully}.

For homophilous graph datasets, we used a constant weight of 1 for \method shown in \cref{lrn formula}. For heterophilic datasets, we used trainable weights instead.
\subsubsection{Hyperbolic SGFoermer Hyperparameters}
We performed a grid search with a constant curvature of -1.0: dimensions within $\{32, 64\}$ for {\sc Chameleon} and {\sc Squirrel}, fixed dimension of 96 for {\sc Film}; learning rate within $\{0.001, 0.01\}$; weight-decay rate within $\{0, 1e-4, 1e-3\}$; dropout rate within $\{0.3, 0.4, 0.5, 0.6\}$; number of layers within $\{2, 3, 4, 5,6, 7, 8\}$.

Here the number of layers refers to the GCN layer utilized in~\cite{wu2023sgformer}. 
\subsection{Implementation details for image datasets}
We performed a grid search where learning rate is within the set $\{0.1, 0.01, 0.001\}$ and weight decay is within the set $\{0, 5e-4, 5e-3\}$. For OOD-detection, we trained the model with the same hyperparameter as in~\cite{van2023poincar} and using the Riemmanian Adam optimizer as in~\cite{van2023poincar} as well. In both cases, we performed search for curvature in the set $\{0.1, 1, 1.5\}$ and the scaling coefficient in \cref{scaling formula} in the set $\{0.5, 1.0, 2.0, \mathrm{trainable}\}$. 

\subsection{Further details and code}
All experiments were performed on float32 datatype and on a single NVIDIA RTX 3070 GPU with 8gb of memory. For details on our implementations, please refer to the code at the \href{https://github.com/Graph-and-Geometric-Learning/LResNet}{\textcolor{blue!50!black}{GitHub link}}.

\subsection{Analysis of curvature as hyperparameter}\label{curv_hyp}
We examine the sensitivity of \method to the choice of curvature $K$ of the hyperbolic manifold by conducting the link prediction and node classification experiments in \cref{GNNs Exp} for $K = -0.5, -1.0, -1.5, -2.0, \mathrm{trainable}$ on the {\sc{airport}} and {\sc{disease}} datasets. The results are shown in \cref{many_curv}. We do not observe the model to be sensitive to curvature choices.

\end{document}